%% file: main.tex
\documentclass[10pt,a5paper,headings=small,listof=totoc,headsepline,DIV=calc,BCOR=12mm,captions=tableheading,fleqn,runningheads,final]{ocmproc}

\usepackage{etex}
\usepackage{ifthen}
\usepackage{makeidx}
\usepackage[english,ngerman]{babel}
\usepackage{graphicx}
\usepackage[utf8]{inputenc}
\usepackage{tabularx}
\usepackage{amsfonts}
\usepackage[mathscr]{eucal}
\usepackage{listings}
\usepackage[font=footnotesize,labelfont=bf,format=hang]{caption} 
\usepackage{color}
\usepackage[table]{xcolor}
\usepackage[standard,thmmarks]{ntheorem}
\usepackage{url}
\usepackage[amssymb]{SIunits}
\usepackage{bibunits}
\usepackage{amsmath}
\usepackage[footnotesize,bf,hang]{subfigure} 
\usepackage{cite}
\usepackage{nicefrac}
\usepackage{dsfont}
\usepackage{enumerate}
\usepackage{wrapfig}
\usepackage{emptypage}
\usepackage{xspace}
\usepackage{textcomp}
\usepackage{tikz}
\usetikzlibrary{matrix,arrows.meta}
\usepackage{wasysym}

\usepackage[sc]{mathpazo}    
\usepackage[scaled]{helvet}  

\DeclareFontEncoding{LGR}{}{}
\DeclareTextSymbol{\~}{LGR}{126}

\newenvironment{itemize*}%
  {\begin{itemize}%
    \setlength{\itemsep}{1pt}%
    \setlength{\parskip}{1pt}}%
  {\end{itemize}}

\newenvironment{enumerate*}%
  {\begin{enumerate}%
    \setlength{\itemsep}{1pt}%
    \setlength{\parskip}{1pt}}%
  {\end{enumerate}}

\newenvironment{description*}%
  {\begin{description}%
    \setlength{\itemsep}{1pt}%
    \setlength{\parskip}{3pt}}%
  {\end{description}}


\begin{document}

	\selectlanguage{english}
  
	\mainmatter

	\begin{bibunit}[IEEEtran]
		\input{content.tex}

		\putbib[content]
	\end{bibunit}
	\cleardoublepage

	\renewcommand{\bibname}{References}
	\cleardoublepage

\end{document}

%% file: content.tex
\newcommand{\h}[2]{{{}^{#1}H_{#2}}}
\newcommand{\p}[2]{{{}^{#1}P_{#2}}}
\newcommand{\FrameScriptStyle}{\scriptscriptstyle}
\newcommand{\FrameLabel}[1]{%
  \mathchoice{\mathrm{#1}}{\mathrm{#1}}%
             {\FrameScriptStyle\mathrm{#1}}%
             {\FrameScriptStyle\mathrm{#1}}%
}
\newcommand{\TCS}{\FrameLabel{TCS}}
\newcommand{\RCS}{\FrameLabel{RCS}}
\newcommand{\CCS}{\FrameLabel{CCS}}
\newcommand{\ACS}[1]{\FrameLabel{ACS}_{#1}}
\newcommand{\PCS}[1]{\FrameLabel{PCS}_{#1}}
\renewcommand{\email}[1]{\texttt{#1}}
\newcommand{\Unit}[1]{\ensuremath{\thinspace\mathrm{#1}}}
\selectlanguage{english}
\title{Laser-Tracker-Assisted Camera-to-Robot Calibration for Mobile Robots}
\titlerunning{Camera-to-Robot Calibration}
\author{Jan A. Rudolph \and \"Oyk\"u Kandemir \and Markus Ulrich}
\authorrunning{J. A. Rudolph et al.}
\tocauthor{J. A. Rudolph, \"O. Kandemir, M. Ulrich}
\institute{Institute of Photogrammetry and Remote Sensing (IPF),\\
Karlsruhe Institute of Technology (KIT),\\
Englerstr. 7, 76131 Karlsruhe, Germany\\
\email{\{jan.rudolph, markus.ulrich\}@kit.edu}\\
\email{oeykue.kandemir@student.kit.edu}}

\abstract{We present a laser-tracker-assisted hand--eye calibration method for camera-equipped mobile robots. The method combines laser-tracker-based 3D metrology with camera-based 2D observations. Building on our previous laser-tracker-assisted camera-to-robot calibration method for ground-observing mobile robots, we present a generalized formulation for calibrating the camera pose in the coordinate system of tracker-localized mobile robots. The new approach relaxes assumptions of our previous method on robot and camera configuration by chaining multiple calibration targets resulting in a more general approach supporting various camera-equipped mobile robot systems.}
\maketitle

\begin{figure}
    \centering
    \includegraphics[width=0.67\linewidth]{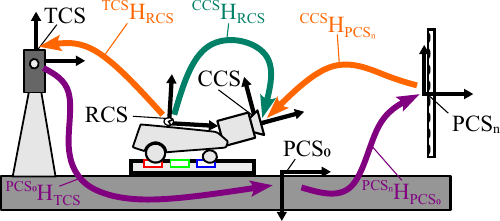}
    \caption{\textbf{Coordinate systems and transformations} relevant for camera-to-robot calibration, extending \cite{RudolphHaitzUlrich2026b}: tracker (\(\TCS\)), robot (\(\RCS\)) and calibration plates (\(\PCS{i}\)). The robot-mounted camera (\(\CCS\)) observes \(\PCS{n}\).  The reference plate (\(\PCS{0}\)) can be absent in the calibration step. Purple denotes known transformations, orange measurements, and green the result \(\h{\RCS}{\CCS}\).}
    \label{fig:Rudolph:idea}

\end{figure}

\section{Motivation}

This work is based on the mobile robot RITA \cite{RudolphHaitzUlrich2026b}, designed for high-accuracy positioning and measurement tasks. RITA is intended for operation in large indoor industrial facilities, where it navigates on planar factory floors while performing positioning and measurement tasks. A camera integrated in the mobile robot supports accurate 3D reconstruction and localization, requiring reliable and accurate calibration of its pose with respect to the robot coordinate system. For this, we combine external 3D metrology from a laser tracker with camera-based target observations to solve the camera-to-robot calibration problem for tracker-localized mobile robots. This setting is challenging because the camera coordinate system is not directly observable by the laser tracker, and the motion of mobile robots often does not provide the pose diversity required by conventional hand--eye calibration approaches.

Hand--eye calibration is well established for vision-guided industrial robots. Typically, these approaches assume sufficiently diverse calibration motions to estimate the rigid camera-to-robot transformation (i.e., the hand--eye pose) unambiguously and accurately \cite{LenzTsai1988CVPR,TsaiLenz1989,ParkMartin1994,Daniilidis1999IJRR,Steger2018Book}. In contrast, mobile robots are often restricted in their motion, e.g., in the case of RITA an approximately planar motion. This makes the direct application of conventional hand--eye calibration methods difficult, since the available robot poses do not sufficiently constrain all components of the camera-to-robot transformation. A similar challenge occurs in the hand-eye calibration of SCARA robots, whose rotational joints are all parallel. In this case, the hand-eye pose is only observable up to an unknown translation along the rotation axes. Although this ambiguity can be resolved by moving the tool to a predefined height \cite{UlrichSteger2016PRIA}, this approach is not applicable to the mobile robot, as it would require vertical motion and physical access to the center of the robot coordinate system.  The present work addresses this gap through a metrology-supported target-registration workflow.

This work builds on our previous laser-tracker-assisted camera-to-robot calibration method for ground-observing mobile robots \cite{RudolphHaitzUlrich2026b}. The method utilizes a dedicated referencing plate that combines reflector nests for laser-tracker-based pose estimation with an integrated camera calibration target. The present work generalizes this plate-based approach beyond the ground-observing setup. Instead of relying on an in-house-designed drive-on referencing plate, the proposed method successively links multiple calibration targets. The chain removes the need for a dedicated combined plate; we nevertheless reuse the existing metrologically characterized reference plate as a convenient tracker anchor. Previous work on hand--eye calibration and calibration uncertainty from both machine-vision and geodetic perspectives \cite{UlrichHillemann2021ICRA,UlrichHillemann2024TRO,Ulrich2024ISPRS} is also relevant in this context, especially \cite{UlrichSteger2016PRIA}, which dealt with restricted pose diversity, exploiting a SCARA-specific geometric constraint. In contrast, we focus on a practical calibration procedure for camera-equipped mobile robots localized by a laser tracker rather than on manipulator kinematics or SCARA-like configurations.

The contribution of this work is a generalized laser-tracker-assisted camera-to-robot calibration workflow that extends our previous work \cite{RudolphHaitzUlrich2026b}, which only supported a nadir-camera setup, to more general camera mounting configurations. The method combines the laser tracker as an external metric reference with image-based observations of calibration targets. It introduces a chained target-registration procedure that transfers the spatial relation from the camera-observed calibration target to the reflector nests. We evaluate the approach on a real mobile robot by assessing the precision of the estimated camera-to-robot transformation and determining its absolute accuracy through a 3D measurement task. The results are compared against ground-truth measurements obtained manually using a laser tracker.

\section{Method}
Our proposed method generalizes our previous work \cite{RudolphHaitzUlrich2026b}, which relied on a dedicated reference plate for calibration that incorporates reflector nests, which support the reflectors of the laser tracker, and a camera calibration target on a common plane. The nadir camera directly observed the reference plate. Instead, our new approach does not require a nadir camera but also works with cameras in arbitrary orientations (for an example of a robot with a front-looking camera see Fig.~\ref{fig:Rudolph:idea}). This is achieved by distinguishing between the reference plate and the plate that is observed by the camera yielding one calibration object. This allows to place the observed plate such that it can be acquired with the robot's camera.

Let $\h{A}{B}$ be the homogeneous rigid transformation that maps 3D points from coordinate system $B$ to $A$, and $\p{A}{j}$ a homogeneous 3D point $j$ expressed in $A$. We distinguish the tracker coordinate system (\(\TCS\)), robot coordinate system (\(\RCS\)), coordinate system of the robot-mounted camera (\(\CCS\)), auxiliary-camera coordinate system in image $i$ (\(\ACS{i}\)), and coordinate system of calibration plate $i$ (\(\PCS{i}\)) (see Fig.~\ref{fig:Rudolph:idea}).

\subsection{Calibration Object}
\label{sec:Rudolph:calibration_object}

\begin{figure}[htb]
\centering
\includegraphics[width=.49\linewidth,trim=0 9.6bp 0 43.2bp,clip]{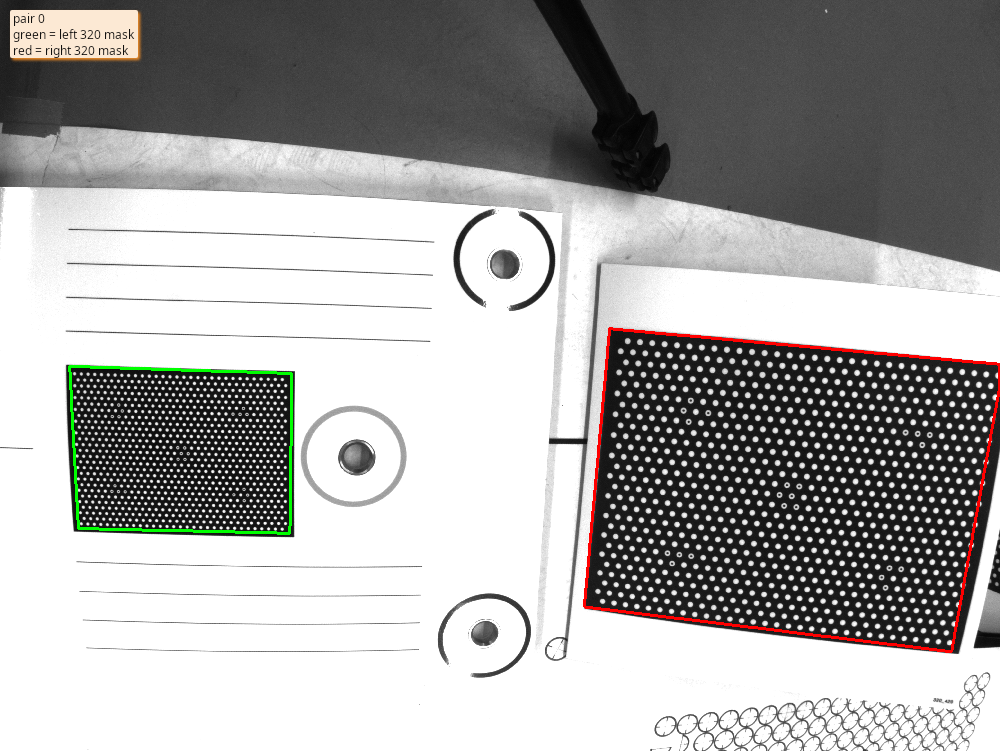}\hfill
\includegraphics[width=.49\linewidth,trim=0 9.6bp 0 43.2bp,clip]{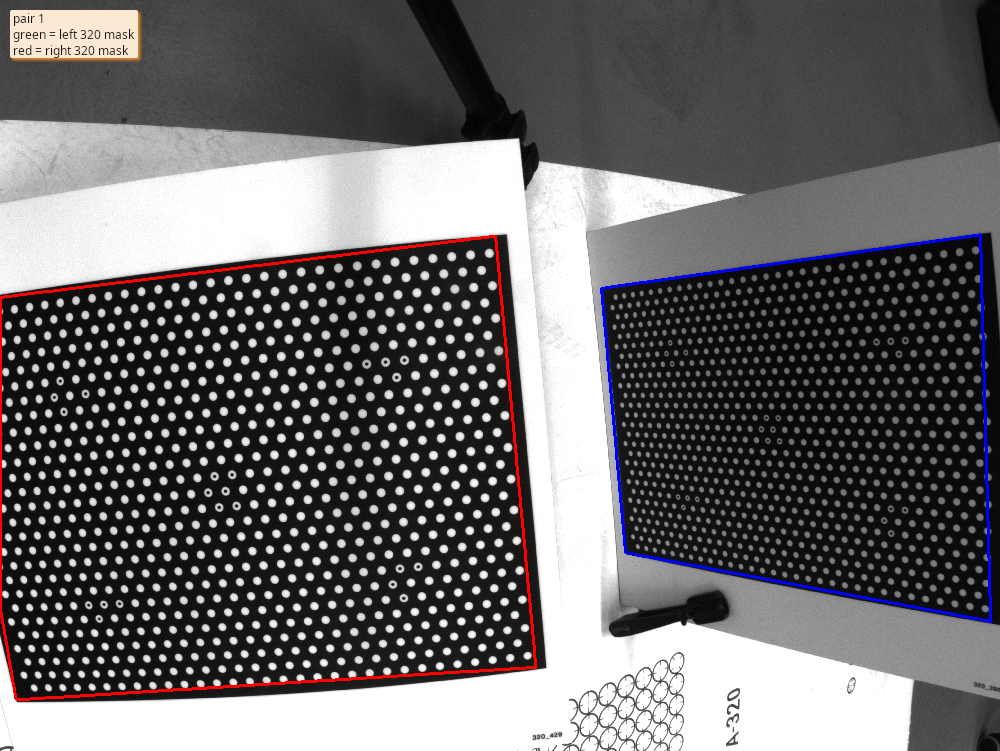}
\caption{Two successive auxiliary camera images with detected HALCON \cite{MVTecHALCON-misc} calibration targets: \(\PCS{0}\) (green) and \(\PCS{1}\) (red) in image 1; \(\PCS{1}\) (red) and \(\PCS{2}\) (blue) in image 2. The shared target links the observations. Images are cropped for visualization.}
\label{fig:Rudolph:plate-separation-example}
\end{figure}

To create the calibration object, the relative pose of the observed plate with respect to the reference plate must be determined prior to hand-eye calibration by taking an image showing both plates with a calibrated auxiliary camera. If both plates cannot be imaged in a single image, alternatively, a chain of calibration plates can be linked by multiple images (see Fig.~\ref{fig:Rudolph:plate-separation-example}). Adjacent images in this chain must share a common fixed target. Note that because these images contain two calibration targets, both calibration targets must be distinguished and identified correctly during the automatic detection. 

An auxiliary camera acquires $n$ images from different poses with coordinate systems \(\ACS{i}\), $i=1,\ldots,n$. Image $i$ shows the adjacent calibration plates \(\PCS{i-1}\) and \(\PCS{i}\). Combining the two target poses detected in image $i$ yields $\h{\PCS{i}}{\PCS{i-1}}=(\h{\ACS{i}}{\PCS{i}})^{-1}(\h{\ACS{i}}{\PCS{i-1}})$. The complete image chain connects the reference plate \(\PCS{0}\) to the plate  \(\PCS{n}\) that is observed by the robot during calibration: $\h{\PCS{n}}{\PCS{0}}=\overleftarrow{\prod}_{i=1}^{n}\h{\PCS{i}}{\PCS{i-1}}$ (the arrow indicates multiplications from the left).

\begin{figure}[htb]
    \centering
    \includegraphics[width=0.55\linewidth]{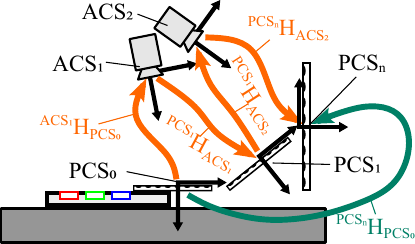}
    \caption{Example target chain with $n=2$ auxiliary camera poses. \(\PCS{0}\) is the reference plate, \(\PCS{1}\) the intermediate plate, and \(\PCS{n}=\PCS{2}\) the plate observed by the robot. Auxiliary camera 1 observes \(\PCS{0}\) and \(\PCS{1}\), auxiliary camera 2 \(\PCS{1}\) and \(\PCS{2}\). Observed poses are displayed in orange, the resulting pose in green.}
    \label{fig:Rudolph:fortpflanzung}
\end{figure}

As described in \cite{RudolphHaitzUlrich2026b} in more detail, a stereo measurement yields the coordinates of the 3D centers of three circular reflector nests $\p{\PCS{0}}{j}$, $j\in\{r,g,b\}$ (red, green, blue) relative to the reference calibration plate, while also incorporating an offset of the distance to the centers of the inserted reflectors. The centers of the three reflectors $\p{\TCS}{j}$ are measured by the laser tracker and used together with $\p{\PCS{0}}{j}$ to estimate $\h{\PCS{0}}{\TCS}$ \cite[Eq.~(13)]{RudolphHaitzUlrich2026b}.

\begin{figure}
    \centering
    \includegraphics[width=0.65\linewidth]{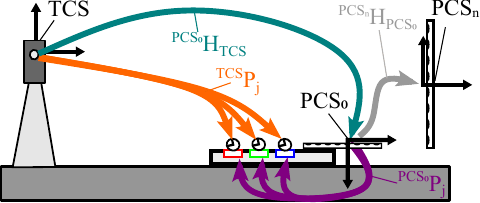}
    \caption{Reference-plate registration \cite[Eq.~(13)]{RudolphHaitzUlrich2026b}. The three laser tracker measured reflector centers \(\p{\TCS}{j}\) (orange) are matched to the known \(\p{\PCS{0}}{j}\) (purple), yielding \(\h{\PCS{0}}{\TCS}\) (green). Note that the robot-observed plate (\(\PCS{n}\)) is not registered directly against the laser tracker measurements.}
    \label{fig:Rudolph:platepose}
\end{figure}

This process yields the calibration object, which can be used to calibrate a series of robots.

\subsection{Camera--Robot Calibration}
\label{sec:Rudolph:calibration}

To calibrate the robot, the robot-mounted camera acquires an image of the calibration target ($\PCS{n}$), which yields $\h{\CCS}{\PCS{n}}$, and hence, $\h{\CCS}{\PCS{0}}=\h{\CCS}{\PCS{n}}\h{\PCS{n}}{\PCS{0}}$.
The robot pose $\h{\TCS}{\RCS}$ is obtained from the laser tracker, which measures the position of the reflector mounted on the robot, and from inertial measurement information. Alternatively, the rotation part of $\h{\TCS}{\RCS}$ is obtained by the method described in \cite[Eqs.~(14)--(21)]{RudolphHaitzUlrich2026b}.  Finally, the camera-to-robot pose is composed as
\begin{align}
 \h{\CCS}{\RCS}&=\h{\CCS}{\PCS{0}}\h{\PCS{0}}{\TCS}\h{\TCS}{\RCS}.
 \label{eq:Rudolph:method:extrinsic}
\end{align}

\section{Results}

\begin{figure}[tbp]
    \centering
    \subfigure[]{\includegraphics[width=\linewidth]{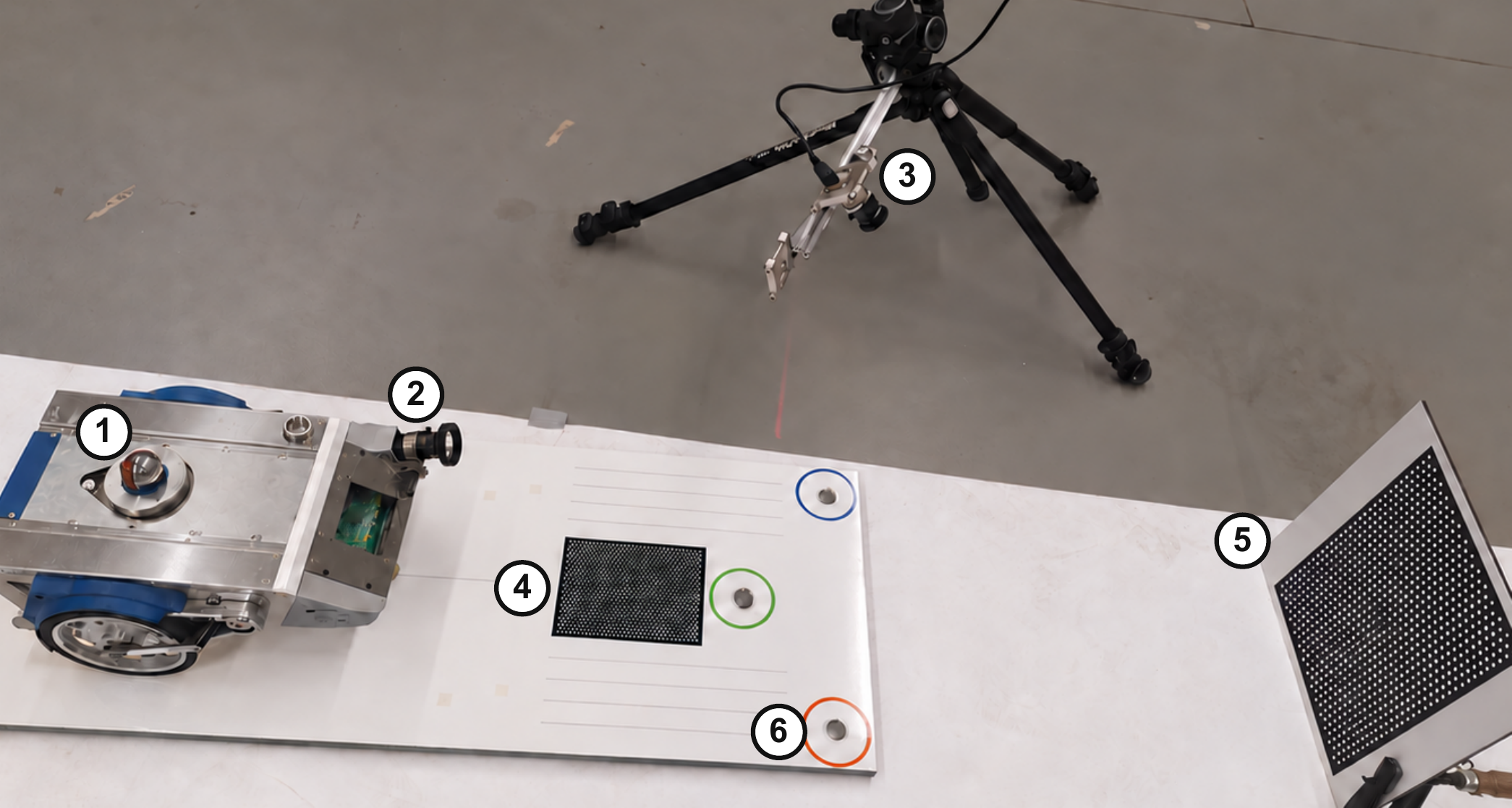}}\\[-0.5ex]
    \subfigure[]{\includegraphics[height=0.275\linewidth]{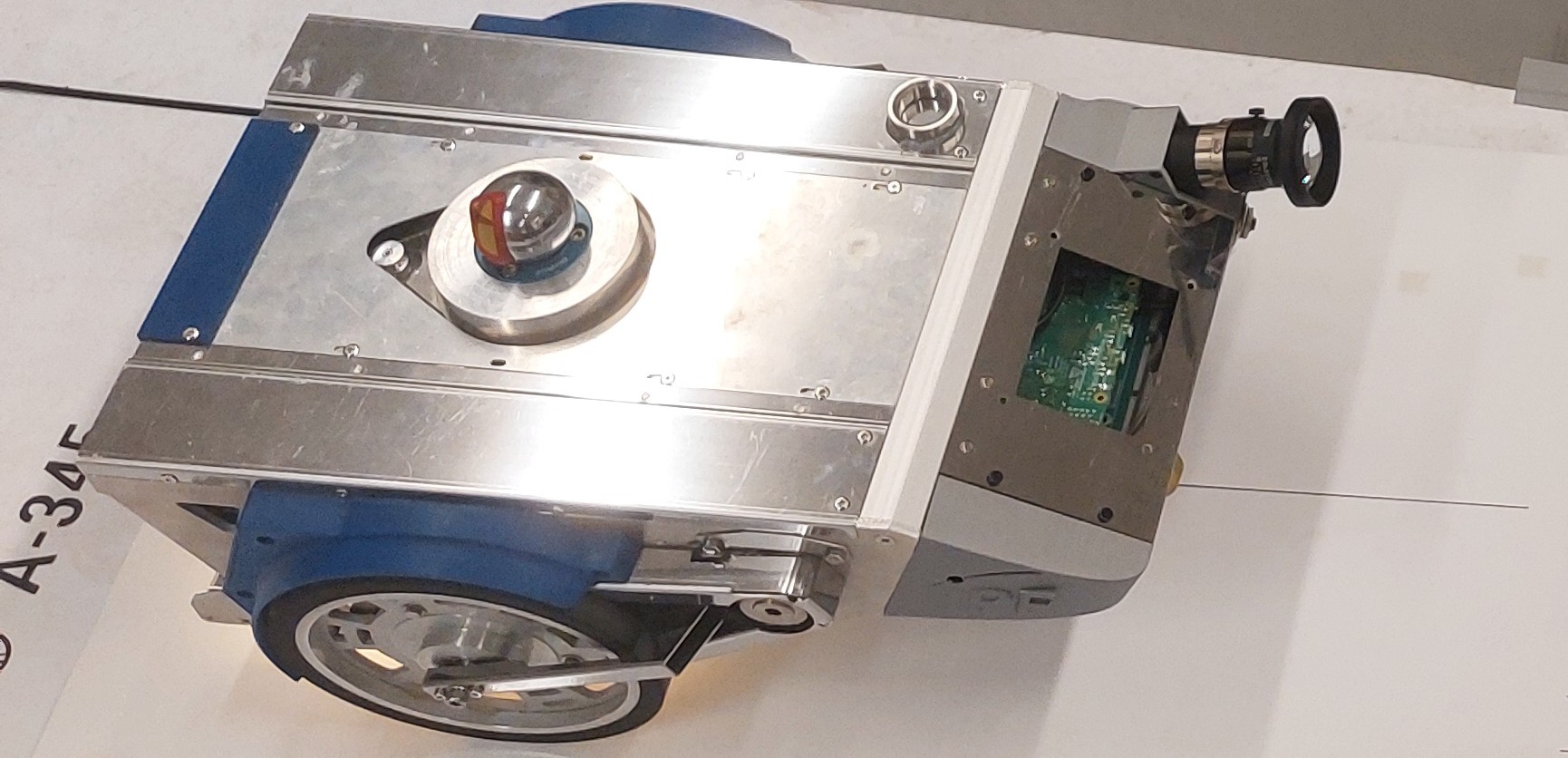}}\hfill
    \subfigure[]{\includegraphics[height=0.275\linewidth]{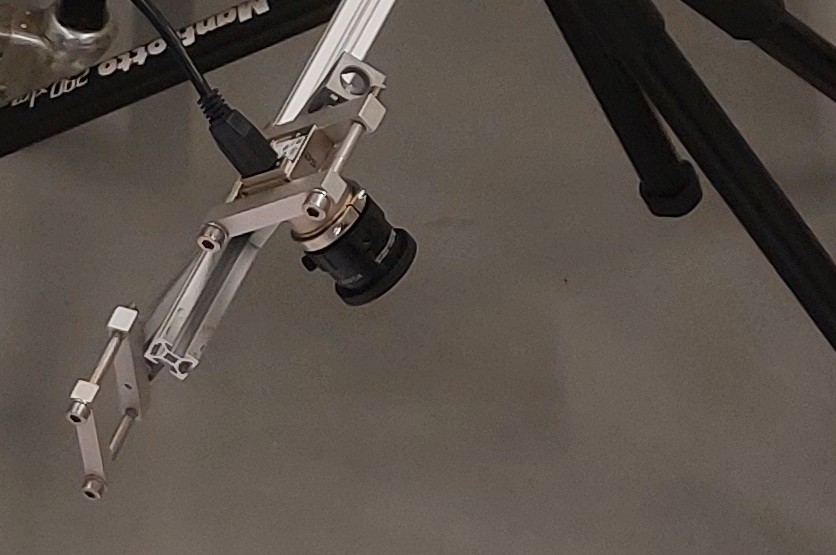}}
    \caption{\textbf{Experimental setup.} (a) and (b): the mobile robot guided by a Leica Absolute Tracker AT901 \cite{leicageosystemsLeicaAbsoluteTracker}, with a Leica/Hexagon 1.5-inch break-resistant corner-cube reflector (1), and the robot-mounted high-resolution imaging system (2): IDS camera (5328 $\times$ 4608 pixels, $2.74\Unit{\mu m}$ pixel pitch) \cite{IDS-U3-31R0CP} with a Schneider-Kreuznach 12\Unit{mm} lens \cite{SchneiderTOPAZ2812}. (a) and (c): the auxiliary camera system (3): IDS camera (5136 $\times$ 3856 pixels, $1.40\Unit{\mu m}$ pixel pitch) \cite{IDS-U3-36P0XCP} with a Schneider-Kreuznach 4.8\Unit{mm} lens \cite{SchneiderCITRINE1848}. Furthermore, (a) shows the HALCON \cite{MVTecHALCON-misc} calibration targets: the reference plate (\(\PCS{0}\)) is a 160\Unit{mm} target (4), the robot-observed plate (\(\PCS{n}\)) is a 320\Unit{mm} target (5). Representative of the three nests \((r,g,b)\) described in Sect.~\ref{sec:Rudolph:calibration_object}, (6) identifies the red reflector nest \(r\). Image (a) shows an exemplarily calibration setup in Scene A.}
    \label{fig:Rudolph:experimental-setup}
\end{figure}

In a first experiment, we evaluate the precision of the camera-to-robot pose, in the second experiment, its absolute accuracy. In our previous work with the nadir setup \cite{RudolphHaitzUlrich2026b}, repeated measurements of a printed floor mark were used for verification. Consequently, only the precision could be evaluated. Because the outward-facing camera observes a larger workspace and permits measuring reference points with a laser tracker, now we are able to evaluate the absolute accuracy as well. Fig.~\ref{fig:Rudolph:experimental-setup} shows the actual hardware that was used and its specifications.

\newcommand{\FrontNStored}{256}
\newcommand{\FrontNExcluded}{16}
\newcommand{\FrontNValid}{240}
\newcommand{\FrontNImagesStored}{32}
\newcommand{\FrontNImagesExcluded}{2}
\newcommand{\FrontNImagesValid}{30}
\newcommand{\FrontNPZero}{128}
\newcommand{\FrontNPOne}{112}
\newcommand{\FrontNVariantA}{120}
\newcommand{\FrontNVariantB}{120}
\newcommand{\FrontNReferences}{eight}
\newcommand{\FrontNRuns}{16}
\newcommand{\FrontSceneARegistrationRmsMm}{0.111522}
\newcommand{\FrontSceneBRegistrationRmsMm}{0.105900}
\newcommand{\FrontNAtMostOneCm}{122}
\newcommand{\FrontPercentAtMostOneCm}{50.83}
\newcommand{\FrontMedianDistanceCm}{0.957}
\newcommand{\FrontMeanDistanceCm}{1.455}
\newcommand{\FrontRmsDistanceCm}{1.755}
\newcommand{\FrontMaxDistanceCm}{3.824}
\newcommand{\FrontMedianPixel}{28.239}
\newcommand{\FrontMaxPixel}{116.794}
\newcommand{\FrontSceneMeanDistanceMm}{4.223}
\newcommand{\FrontSceneMeanAngleDeg}{0.643}

\subsection{Precision}
\label{sec:Rudolph:evaluation-calibration}

To evaluate the precision of our approach, we perform the creation of the calibration object (Sect.~\ref{sec:Rudolph:calibration_object}) and the camera--robot calibration (Sect.~\ref{sec:Rudolph:calibration}) \FrontNReferences{} times. In the first $N=4$ experiments (Scene A), we used a single intermediate calibration plate (i.e., $n=2$) and varied the poses of the intermediate and the observed plate. We chose a distance between reference and observed plate of about 0.9\Unit{m}. In the second $N=4$ experiments (Scene B) we rotated the reference plate by approx. 180 degrees about the vertical axis and used two intermediate calibration plates (i.e., $n=3$) to cover the larger distance between reference and observed plate of about 1.3\Unit{m}. Consequently, Scene A uses two chain images, whereas Scene B uses three to span the larger distance. 

\begin{figure}[tbp]
 \centering
 \includegraphics[width=\linewidth]{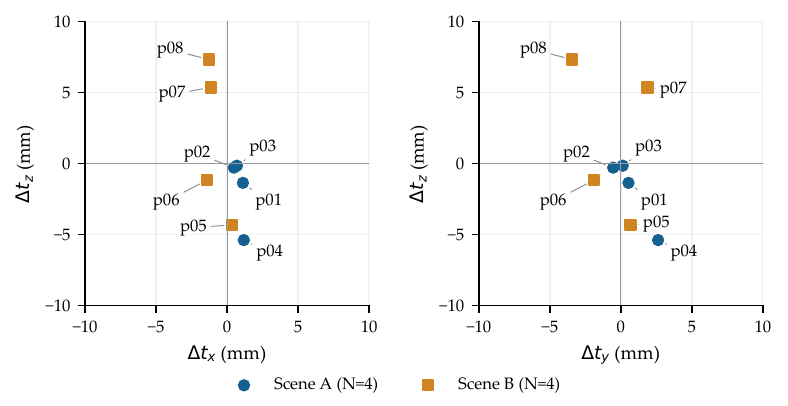}
 \caption{\textbf{Results of precision evaluation.} Deviation of the translation component of $\h{\CCS}{\RCS}$ with respect to the mean of the \FrontNReferences{} camera-to-robot estimates. Blue circles are calibration results of Scene A, orange squares results of Scene B.}
 \label{fig:Rudolph:calibration-scatter}
\end{figure}

Table~\ref{tab:Rudolph:calibrationmetrics} and Figure~\ref{fig:Rudolph:calibration-scatter} show the results of the precision evaluation for Scene A and B. The lower precision of the results of Scene B are obvious and are caused by the larger distance of the reference plate from the observed plate, which made an additional intermediate calibration target pose necessary. While for our localization application, the precision is sufficient, in general the calibration chain length should be minimized to avoid error accumulation. From Figure~\ref{fig:Rudolph:calibration-scatter}, one can see that the precision in z direction (i.e., the viewing direction of the camera) is worse than in lateral direction.


\begin{table}[tbp]
\centering
\setlength{\tabcolsep}{3pt}
\begin{tabular}{lrrrrr}
\hline
 & & \multicolumn{2}{c}{RMS to scene mean} & \multicolumn{2}{c}{Max. pair distance} \\ Group & $N$ & mm & deg & mm & deg \\
\hline
Scene A & 4 & 2.449 & 0.124 & 6.041 & 0.288 \\
Scene B & 4 & 5.216 & 0.360 & 12.457 & 0.717 \\
\hline
\end{tabular}
\par\smallskip
\footnotesize
\caption{\textbf{Results of precision evaluation.} RMS values of the translation and rotation component of $\h{\CCS}{\RCS}$ with respect to the average translation vector and average rotation matrix. Maximum differences between pairs of experiments in translation and rotation.}\label{tab:Rudolph:calibrationmetrics}
\end{table}

\subsection{Accuracy}
\label{sec:Rudolph:evaluation-point}

\begin{figure}[tbh]
 \centering
 \includegraphics[width=\linewidth]{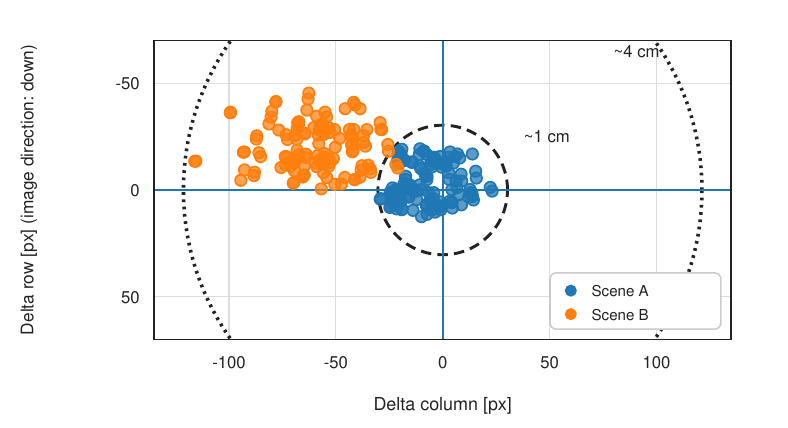}
 \vspace{-0.5cm}
 \caption{\textbf{Results of accuracy evaluation.} Offset of the measured image points to the reprojected ground truth of \FrontNValid{} experiments, distinguishing experiments of Scene A (blue) and B (orange). For visualization purposes, circles of points with equidistant offsets of 1\Unit{cm} and 4\Unit{cm} are projected into the image.}
 \label{fig:Rudolph:offset-scatter}
\end{figure}

To evaluate the absolute accuracy of the camera--robot pose, the ground truth 3D coordinates of a marked reference point are measured with a laser tracker. Then, an image of the reference point is acquired with the robot's camera and its image coordinates are determined. This image point defines a line-of-sight in 3D space. As evaluation metrics we measure the orthogonal distance of the 3D reference point from the line-of-sight as well as its reprojection error.

To transform measurements from $\CCS$ into $\TCS$, the robot pose is obtained from laser tracker positions and IMU orientation. Because of the planar robot motion, we assume zero roll and pitch (see \cite[Eqs.~(14)--(21)]{RudolphHaitzUlrich2026b}). With the inverse robot pose $\h{\RCS}{\TCS}$, the calibration result $\h{\CCS}{\RCS}$ yields the pose $\h{\CCS}{\TCS}=\h{\CCS}{\RCS}\h{\RCS}{\TCS}$ of the camera in the $\TCS$.

The measured image point $\mathbf u=(row,col)^\top$ and the calibrated camera model define the line-of-sight in the $\CCS$. Because the line-of-sight passes through the origin of the $\CCS$ it can be described by the vector ${}^{\CCS}\mathbf d$. The tracker-measured ground truth reference point ${}^{\TCS}P_g$ is transformed into the $\CCS$ by ${}^{\CCS}\bar P_g=\h{\CCS}{\TCS}\,{}^{\TCS}\bar P_g$ (where $\bar P$ denote homogeneous coordinates of $P$). Note that $\h{\CCS}{\TCS}$ contains the calibrated camera--robot pose that is to be evaluated. Then, the orthogonal distance of the reference point to the line-of-sight is
\begin{align}
  e_\perp&=\left\lVert{}^{\CCS}P_g-
 [({}^{\CCS}\mathbf d)^\top{}^{\CCS}P_g]{}^{\CCS}\mathbf d\right\rVert.
 \label{eq:Rudolph:method:line-projection}
\end{align}
\begin{samepage}
The reprojection error is computed as
\begin{align}
 \Delta\mathbf u&=\pi({}^{\CCS}P_g)-\mathbf u
 =(\Delta row,\Delta col)^\top,\qquad
 e_{px}=\lVert\Delta\mathbf u\rVert,
 \label{eq:Rudolph:method:pixel}
\end{align}
\end{samepage}
where $\pi({}^{\CCS}P_g)$ is the 3D reference point projected into the image by using the calibrated camera model.

For the evaluation, we moved the robot to \FrontNImagesValid{} different poses and took an image of the reference point from each pose. Then, for all calibration results of the eight experiments of Scene A and B from Sect.~\ref{sec:Rudolph:evaluation-calibration} and for each image, we computed both evaluation metrics. Table~\ref{tab:Rudolph:pointmetrics-scenes} and Fig.~\ref{fig:Rudolph:offset-scatter} summarize the results. Note that the accuracy of all 120 results of the calibrations using Scene A is better than 1\unit{cm}. As already seen from the precision evaluation, the calibrations using Scene B lead to worse results. Fig.~\ref{fig:Rudolph:offset-scatter} clearly shows that the results of Scene A are scattered around the ground truth while those of Scene B show a significant systematic error.


\begin{table}[tbh]
\centering
\footnotesize
\setlength{\tabcolsep}{3pt}
\begin{tabular}{lr lrrrr}
\hline
Group & Metric & Mean & RMS & Max. \\
\hline
Scene A & $e_\perp$ (cm) & 0.580 & 0.615 & 0.961 \\
Scene A & $e_{px}$ (px) & 16.098 & 17.265 & 29.561 \\
Scene B & $e_\perp$ (cm) & 2.331 & 2.405 & 3.824 \\
Scene B & $e_{px}$ (px) & 64.190 & 66.708 & 116.794 \\
\hline
\end{tabular}
\par\smallskip
\begin{minipage}{\linewidth}\footnotesize
\caption{\textbf{Results of accuracy evaluation.} Absolute errors for Scene A and B, each containing 120 experiments. The metrics $e_\perp$ and $e_{px}$ denote the point-to-line and pixel errors.}
\label{tab:Rudolph:pointmetrics-scenes}
\end{minipage}
\end{table}

\section{Conclusion}
We presented a laser-tracker-assisted camera-to-robot calibration method that extends our previous nadir-camera approach to arbitrarily oriented cameras by connecting the tracker-referenced and robot-observed calibration plates through a chain of calibration targets. The experiments demonstrate the practical applicability of the method and also show that the chain design is critical: the precision and accuracy of all 120 Scene A evaluations are better than 1\Unit{cm}, whereas the longer chain of Scene B yielded a lower precision and the accuracy results show a systematic offset. Thus, the distance between the reference and observed plates and the number of intermediate calibration targets should be minimized to limit error accumulation. Future work will isolate and quantify the individual error contributions: e.g., translation and orientation components of the robot poses, the calibration target poses, and the stability of the setup. This will help to optimize the calibration target arrangements and the methodology, and hence, to further improve accuracy and precision.

\section*{Acknowledgements}
This Project is supported by the Federal Ministry for Economic Affairs and Climate Action (BMWK) on the basis of a decision by the German Bundestag.

The authors used AI tools for language refinement and generation of scripts for statistical analysis and plotting. All content was verified by the authors, who take full responsibility for the final manuscript.


%% file: main.bbl
\begin{thebibliography}{10}
\providecommand{\url}[1]{#1}
\csname url@samestyle\endcsname
\providecommand{\newblock}{\relax}
\providecommand{\bibinfo}[2]{#2}
\providecommand{\BIBentrySTDinterwordspacing}{\spaceskip=0pt\relax}
\providecommand{\BIBentryALTinterwordstretchfactor}{4}
\providecommand{\BIBentryALTinterwordspacing}{\spaceskip=\fontdimen2\font plus
\BIBentryALTinterwordstretchfactor\fontdimen3\font minus \fontdimen4\font\relax}
\providecommand{\BIBforeignlanguage}[2]{{%
\expandafter\ifx\csname l@#1\endcsname\relax
\typeout{** WARNING: IEEEtran.bst: No hyphenation pattern has been}%
\typeout{** loaded for the language `#1'. Using the pattern for}%
\typeout{** the default language instead.}%
\else
\language=\csname l@#1\endcsname
\fi
#2}}
\providecommand{\BIBdecl}{\relax}
\BIBdecl

\bibitem{RudolphHaitzUlrich2026b}
J.~A. Rudolph, D.~Haitz, and M.~Ulrich, ``A novel camera-to-robot calibration method for vision-based floor measurements,'' \emph{ISPRS Annals of the Photogrammetry, Remote Sensing and Spatial Information Sciences}, vol. XI-2-2026, pp. 705--712, 2026.

\bibitem{LenzTsai1988CVPR}
R.~K. Lenz and R.~Y. Tsai, ``Calibrating a cartesian robot with eye-on-hand configuration independent of eye-to-hand relationship,'' in \emph{Proceedings of the CVPR '88: The Computer Society Conference on Computer Vision and Pattern Recognition}, 1988.

\bibitem{TsaiLenz1989}
R.~Y. Tsai and R.~K. Lenz, ``A new technique for fully autonomous and efficient {3D} robotics hand/eye calibration,'' \emph{IEEE Transactions on Robotics and Automation}, vol.~5, no.~3, 1989.

\bibitem{ParkMartin1994}
F.~C. Park and B.~Martin, ``Robot sensor calibration: Solving {AX} = {XB} on the {E}uclidean group,'' \emph{IEEE Transactions on Robotics and Automation}, vol.~10, 1994.

\bibitem{Daniilidis1999IJRR}
K.~Daniilidis, ``Hand-eye calibration using dual quaternions,'' \emph{The International Journal of Robotics Research}, vol.~18, 1999.

\bibitem{Steger2018Book}
C.~Steger, M.~Ulrich, and C.~Wiedemann, \emph{Machine Vision Algorithms and Applications}.\hskip 1em plus 0.5em minus 0.4em\relax Wiley-VCH, 2018.

\bibitem{UlrichSteger2016PRIA}
M.~Ulrich and C.~Steger, ``Hand-eye calibration of {SCARA} robots using dual quaternions,'' \emph{Pattern Recognition and Image Analysis}, vol.~16, no.~1, 2016.

\bibitem{UlrichHillemann2021ICRA}
M.~Ulrich and M.~Hillemann, ``Generic hand--eye calibration of uncertain robots,'' in \emph{2021 IEEE International Conference on Robotics and Automation (ICRA)}, Xi'an, China, 2021.

\bibitem{UlrichHillemann2024TRO}
------, ``Uncertainty-aware hand-eye calibration,'' \emph{IEEE Transactions on Robotics}, vol.~40, 2024.

\bibitem{Ulrich2024ISPRS}
M.~Ulrich, C.~Steger, F.~Butsch, and M.~Liebe, ``Vision-guided robot calibration using photogrammetric methods,'' \emph{ISPRS Journal of Photogrammetry and Remote Sensing}, vol. 218, 2024.

\bibitem{MVTecHALCON-misc}
{MVTec Software GmbH}, ``{MVTec HALCON},'' \url{https://www.mvtec.com/products/halcon}, Munich, Germany, 2026, accessed September 18, 2026.

\bibitem{leicageosystemsLeicaAbsoluteTracker}
{Leica Geosystems}, ``Leica {{Absolute Tracker AT901}} and {{PCMM}} specs,'' 2010.

\bibitem{IDS-U3-31R0CP}
{IDS Imaging Development Systems GmbH}, ``{uEye U3-31R0CP-C-HQ Rev. 2.2} camera data sheet,'' \url{https://en.ids-imaging.com/store/u3-31r0cp-rev-2-2.html}, 2026, manufacturer data sheet and product page, Accessed September 22, 2026.

\bibitem{SchneiderTOPAZ2812}
{Jos. Schneider Optische Werke GmbH}, ``{TOPAZ 2.8/12 C} lens data sheet,'' \url{https://schneiderkreuznach.com/en/industrial-optics/lenses/c-mount-lenses/topaz/f2-8-12mm-c}, 2026, manufacturer data sheet, Accessed September 22, 2026.

\bibitem{IDS-U3-36P0XCP}
{IDS Imaging Development Systems GmbH}, ``{uEye U3-36P0XCP-C-HQ Rev. 1.2} camera data sheet,'' \url{https://en.ids-imaging.com/store/u3-36p0xcp-rev-1-2.html}, 2026, manufacturer data sheet and product page, Accessed September 22, 2026.

\bibitem{SchneiderCITRINE1848}
{Jos. Schneider Optische Werke GmbH}, ``{CITRINE 1.8/4.8 C-S} lens data sheet,'' \url{https://schneiderkreuznach.com/en/industrial-optics/lenses/c-mount-lenses/citrine/f1-8-4-8mm-c-stabilized}, 2026, manufacturer data sheet, Accessed September 22, 2026.

\end{thebibliography}
